\pdfoutput=1

\documentclass[11pt]{article}

\usepackage[]{acl}

\usepackage{times}
\usepackage{latexsym}

\usepackage[T1]{fontenc}

\usepackage[utf8]{inputenc}

\usepackage{microtype}

\usepackage{inconsolata}

\usepackage{graphicx} %
\usepackage{natbib}  %
\usepackage{caption} %
\usepackage{algorithm}
\usepackage{listings}
\usepackage{algorithmic}
\usepackage{amsmath}
\usepackage{booktabs}
\usepackage{multirow}
\usepackage[outdir=./]{epstopdf}
\usepackage{enumitem}
\usepackage{caption}
\usepackage{subcaption}
\usepackage{subfloat}
\usepackage{newfloat}
\usepackage{graphicx}
\usepackage{svg} %
\usepackage[normalem]{ulem}
\usepackage{framed}
\usepackage{mdframed}
\usepackage{xcolor}
\usepackage{lipsum}
\usepackage{float}
\usepackage{hyperref}

\definecolor{shadecolor}{gray}{0.9}

\newlist{todolist}{itemize}{2}
\setlist[todolist]{label=$\square$}
\usepackage{pifont}

\usepackage{array}
\newcolumntype{L}[1]{>{\raggedright\let\newline\\\arraybackslash\hspace{0pt}}m{#1}}
\newcolumntype{C}[1]{>{\centering\let\newline  \\\arraybackslash\hspace{0pt}}m{#1}}
\newcolumntype{R}[1]{>{\raggedleft\let\newline \\\arraybackslash\hspace{0pt}}m{#1}}

\AtBeginDocument{%
  \providecommand\BibTeX{{%
    \normalfont B\kern-0.5em{\scshape i\kern-0.25em b}\kern-0.8em\TeX}}
}

\title{EconAgent: Large Language Model-Empowered Agents\\ for Simulating Macroeconomic Activities}

\author{Nian Li$^1$ \ \ \ Chen Gao$^2$\thanks{Corresponding author.} \ \ \ Mingyu Li$^2$ \ \ \ Yong Li$^2$\footnotemark[1] \ \ \ Qingmin Liao$^1$ \\
$^1$Shenzhen International Graduate School, Tsinghua University \ \ \ $^2$Tsinghua University \\
\small \texttt{linian21@mails.tsinghua.edu.cn} \ \ \ \texttt{chgao96@gmail.com} \ \ \ \texttt{liyong07@tsinghua.edu.cn}
}

\begin{document}
\maketitle
\begin{abstract}
The advent of artificial intelligence has led to a growing emphasis on data-driven modeling in macroeconomics, with agent-based modeling (ABM) emerging as a prominent bottom-up simulation paradigm. In ABM, agents~(\textit{e.g.}, households, firms) interact within a macroeconomic environment, collectively generating market dynamics. 
Existing agent modeling typically employs predetermined rules or learning-based neural networks for decision-making. However, customizing each agent presents significant challenges, complicating the modeling of agent heterogeneity. Additionally, the influence of multi-period market dynamics and multifaceted macroeconomic factors are often overlooked in decision-making processes.
In this work, we introduce \textbf{EconAgent}, a large language model-empowered agent with human-like characteristics for macroeconomic simulation. We first construct a simulation environment that incorporates various market dynamics driven by agents' decisions regarding work and consumption. Through the perception module, we create heterogeneous agents with distinct decision-making mechanisms. Furthermore, we model the impact of macroeconomic trends using a memory module, which allows agents to reflect on past individual experiences and market dynamics.
Simulation experiments show that EconAgent can make realistic decisions, leading to more reasonable macroeconomic phenomena compared to existing rule-based or learning-based agents. Our codes are released at \url{https://github.com/tsinghua-fib-lab/ACL24-EconAgent}.

\end{abstract}

\section{Introduction}
\label{sec::intro}
The advent of artificial intelligence (AI) has indeed transformed the research paradigm in traditional economics~\cite{jorgenson2001information}. In the digital economy era, individual economic behavior can be elaborately recorded and analyzed, leading to a growing emphasis on data-driven modeling in macroeconomics. These data sources may include online consumption, social media activity, and more, allowing economists to better understand individual economic behavior rather than relying solely on traditional macroeconomic indicators~\cite{schorfheide2015real}.
Research in macroeconomics aims to analyze and predict economic variables quantitatively. Early empirical statistical models, such as the Phelps Model~\cite{phelps1967phillips}, and the work of Kydland and Prescott~\cite{kydland1982time}, focused on data-driven analysis and policy outcome prediction but struggled to handle significant shocks. Later, Dynamic Stochastic General Equilibrium (DSGE) models~\cite{christiano2005nominal} were introduced to capture economic dynamics and shocks but assumed a perfect world.

In the last two decades, agent-based modeling (ABM) has emerged as a promising paradigm for simulating macroeconomics from the bottom up, allowing diverse agents to interact without assuming a predetermined equilibrium~\cite{farmer2009economy}. The evolution of ABM in macroeconomics can be divided into two stages. Early models~\cite{tesfatsion2006handbook,brock1998heterogeneous} relied on predetermined rules but made oversimplified assumptions about agent behaviors. Later learning-based models~\cite{trott2021building,zheng2022ai,mi2023taxai}, were trained with large-scale behavioral data. Essentially, these models transform multifaceted economic factors into agent decisions.
However, customizing decision-making mechanisms for each agent presents substantial difficulties. Using customized rules requires extensive expert knowledge and sophisticated model calibration~\cite{windrum2007empirical}. Similarly, employing customized neural networks leads to a sharp increase in network parameters and difficulties in model training~\cite{mi2023taxai}. These difficulties raise the challenge of modeling agent heterogeneity. Additionally, existing models typically focus on the individual situations of the current period, making it difficult to consider past periods, their variations, and multifaceted macroeconomic factors. This complicates the modeling of the influence of macroeconomic trends and dynamics.

Recently, the field of AI has witnessed the rise of large language models (LLMs)~\cite{zhao2023survey}. Leveraging this advancement, LLM-empowered agents have demonstrated capabilities in reasoning, planning, and decision-making. In this work, we design \textbf{EconAgent}, a LLM-empowered agent with human-like characteristics for macroeconomic simulations. We first construct a simulation environment that includes labor and consumption market dynamics driven by agents' decisions on working and consumption, as well as fiscal and monetary policies. Using a perception module that targets agent profiles and mirrors real-world economic situations, we create heterogeneous agents automatically exhibiting different decision-making mechanisms. Additionally, we model the influence of macroeconomic trends with a memory module, enabling agents to reflect on past individual experiences and market dynamics.

In our experiments, classic macroeconomic phenomena, such as inflation in the consumption market and the unemployment rate in the labor market, are reproduced more reasonably compared to traditional rule-based or learning-based agents. We also observe that the EconAgent exhibits human-like decision-making patterns, reflecting the complexities of human economic behavior. These agents demonstrate swift adaptability in response to changes in the internal and external environment.

The contributions of this work can be summarized as follows:
\begin{itemize}[leftmargin=*]
    \item We take the first step to integrate LLMs into the domain of macroeconomic simulations, bridging two seemingly disparate fields into a cohesive research avenue.
    \item We conduct macroeconomic simulations in our constructed environment driven by EconAgent, resulting in the emergence of classic macroeconomic phenomena and regularities.
    \item The results show that our approach not only enhances the realism and depth of macroeconomic simulations but also provides a promising avenue for future research, potentially reshaping how we study and understand the intricacies of macroeconomics.
\end{itemize}
\section{Framework Overview}\label{sec::system}

\begin{figure*}[t]
\centering
    \includegraphics[width=0.8\linewidth]{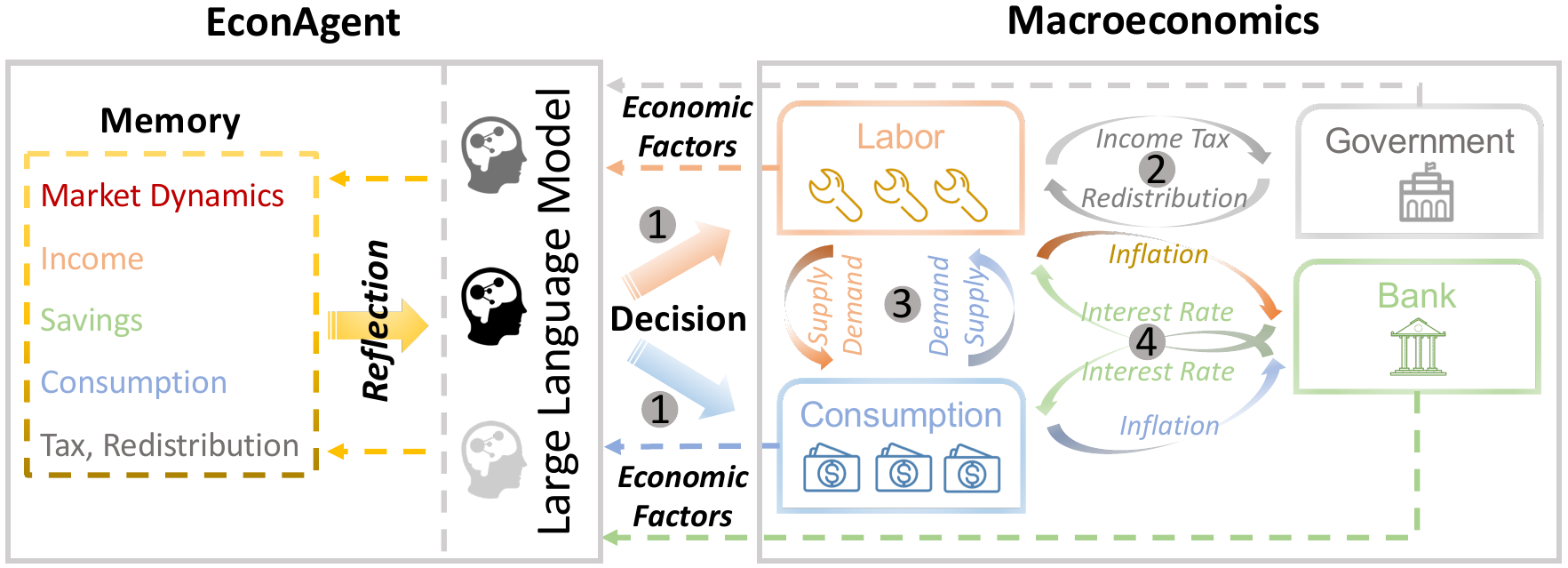}
\caption{The illustration of our EconAgent~(left) and simulation environment~(right).}
\label{fig:framework}
\end{figure*}

In this section, we present the overall framework of the macroeconomic simulation. As illustrated in Figure \ref{fig:framework}, it follows the well-acknowledged simulation frameworks and provides an environment driven by EconAgent. The simulation encompasses four primary components: labor, consumption, financial markets, and government taxation. We simulate the two most critical decisions individuals make in real life: work and consumption~\cite{gatti2011macroeconomics,wolf2013multi,dawid2018agent}, which subsequently influence the fiscal revenues of the government~\cite{zheng2022ai,trott2021building,dawid2018agent} and impact the dynamics of labor and consumption markets~\cite{lengnick2013agent,deissenberg2008eurace,dawid2012eurace}. Accordingly, banks adjust interest rates based on market inflation or deflation~\cite{wolf2013multi,dawid2018agent}.

\subsection{Agent Decisions}
Labor supply and consumption are necessary agent~(household) decisions in macroeconomic simulations. In our framework, each simulation step indicates one month, in which agent $i$ decides,
\begin{itemize}[leftmargin=*]
  
  \item whether to work $l_i\sim Bernoulli(p_i^w)$, where $p_i^w$ is the work propensity. If they decide to work~($l_i=1$), they receive a monthly wage as the income, which varies among agents. Each agent is initialized with an hourly wage $w_i$ following the Pareto distribution~\cite{zheng2022ai}, and the monthly wage $v_i$ is calculated by multiplying $168$ hours~(21 working days at 8 hours/day~\cite{lengnick2013agent}). Those who abstain from work~($l_i=0$) have an income of zero.
  \item the consumption propensity $p_i^c$, indicating the proportion of their wealth (including their current savings and income in this month) they wish to spend for essential goods.

\end{itemize}
As the first challenge, simulating heterogeneous agents' decisions is vital for the emergence of macroeconomic phenomena. Moreover, each agent is influenced by multifaceted economic factors, such as the expected income, the tax paid, \textit{etc}. However, conventional simulations typically model a limited number of factors via predetermined equations~\cite{lengnick2013agent,gatti2011macroeconomics,wolf2013multi}. These models lack of flexibility to simulate diverse decision-making mechanisms and assume one or a few representative rules~\cite{axtell2022agent}. 

\subsection{Government Taxation}
The government assumes the responsibility of taxation and provision of public services in society, as well as fiscal redistribution to ensure social equity. Taxes are collected from all the agents' income\footnote{We only consider income tax in this work and leave other taxes for future work, such as value-added tax.}. The progressive tax for income $z_i$ is calculated as follows,
\begin{align}
    T(z_i) &= \sum_{k=1}^{B} \tau_{k}\left(\left(b_{k+1}-b_{k}\right) \mathbf{1}\left[z_i>b_{k+1}\right]\right. \nonumber \\
    &\left.+\left(z_i-b_{k}\right) \mathbf{1}\left[b_{k}<z_i \leq b_{k+1}\right]\right),
\end{align}
where $b_k$ is the $k$-th tax bracket, $\tau_k$ is corresponding tax rate, and $\mathbf{1}[\cdot]$ is indicator function. The tax brackets and rates are set as the 2018 U.S. Federal tax schedule.

The tax revenue is then evenly redistributed among all the agents. Therefore, the post-tax income is
\begin{equation}
    \hat{z}_i = z_i - T(z_i) + z^r = z_i - T(z_i) + \frac{1}{N}\sum_{j=1}^N T(z_j),
\end{equation}
where $z^r$ indicates the redistribution. The individual savings for the agent are then updated as follows,
\begin{equation}
    s_i \leftarrow s_i + \hat{z}_i
\end{equation}

\subsection{Productivity and Consumption}
Incorporating agent decisions and government taxation, we simulate labor and consumption market dynamics based on economic principles. First, working agents contribute 168 hours of productivity monthly, translating to the production of essential goods\footnote{Note that we leave the simulation of firms as future work.}. The inventory of goods $G$ is then updated as,
\begin{equation}
    G \leftarrow G + S = G + \sum_{j=1}^N l_j\times 168\times A,
\end{equation}
where $S$ denotes the production volume from agents' labor supply, and $A$ is a universal productivity.
As for the consumption, the total demand of goods is,
\begin{equation}
    D = \sum_{j=1}^N d_j = \sum_{j=1}^N \frac{c_j}{P} = \sum_{j=1}^N \frac{p_j^c s_j}{P},
\end{equation}
where $d_j$ is the intended demand of agent $j$, $c_j$ is the intended consumption, $s_j$ is current savings, and $P$ is the price of essential goods. Furthermore, both labor and consumption markets evolve based on the imbalance between supply and demand. Specifically, the imbalance is defined as,
\begin{equation}
    \Bar{\varphi} = \frac{D - G}{\max(D, G)},
\end{equation}
When the essential goods are in shortage, \textit{i.e.}, the supply can not meet the demand, the worker's wage should be increased to stimulate production. Due to the rise in labor costs for firms, they will also increase the goods price to ensure a certain profit margin~\cite{lengnick2013agent,dawid2018agent,wolf2013multi}. The hourly wage is adjusted as follows,
\begin{equation}
    w_i \leftarrow w_i (1 + \varphi_i), \varphi_i \sim sign(\Bar{\varphi}) U(0, \alpha_w\vert\Bar{\varphi}\vert),
\end{equation}
and the price is adjusted as follows,
\begin{equation}
    P \leftarrow P (1 + \varphi_P), \varphi_P \sim sign(\Bar{\varphi})U(0, \alpha_P\vert\Bar{\varphi}\vert),
\end{equation}
where $\alpha_w$ and $\alpha_P$ are the maximum rate of change when adjusting the wage and price, respectively.
We also simulate the dynamics of goods consumption. Specifically, an agent $j$ is randomly selected to consume essential goods, and real consumption goods and money are limited by current inventory of goods,
\begin{equation}
    \hat{d}_j = \min(d_j, G), \hat{c}_j = \hat{d}_j\times P
\end{equation}
which means the demand is met if, and only if, there is a sufficient supply. The inventory of total goods also decreases,
\begin{equation}
    G \leftarrow G - \hat{d}_j.
\end{equation}
The process continues until every agent has consumed goods once.

\subsection{Financial Market}
Annually, the savings of each agent increase based on the interest rate set by the bank,
\begin{equation}
    s_i \leftarrow s_i\times (1 + r).
\end{equation}
Furthermore, in the first month of each year, the bank adjusts the interest rate based on the inflation in the consumption market and the unemployment rate in the labor market. Specifically, we adopt the widely-used Taylor rule to set the interest rate~\cite{wolf2013multi,dawid2018agent},
\begin{equation}
    r = \max(r^n + \pi^t + \alpha^\pi(\pi - \pi^t) + \alpha^u(u^n - u), 0),
\end{equation}
where $r^n$ and $u^n$ indicate the natural interest rate and unemployment rate, respectively. Besides, $\pi^t$ is the target inflation rate. The interest rate is adjusted adaptively according to annual inflation rate $\pi$ and unemployment rate $u$, where $\alpha^\pi$ and $\alpha^u$ denote inflation and unemployment adaptation coefficients, respectively. We define the inflation and unemployment rate as follows,
\begin{equation}
\label{def-indicator}
    \pi = \frac{\Bar{P}_n - \Bar{P}_{n-1}}{\Bar{P}_{n-1}}, u = \frac{\sum_{m=1}^{12}\sum_{j=1}^N (1-l_j)}{12 N},
\end{equation}
where $\Bar{P}_n$ is the average goods price over the $n$-th year, and $m$ denotes the $m$-th month.

When considering the dynamics of labor, consumption, and financial markets, the influence of these macroeconomic trends on agent decision-making is also seldom considered, raising the second challenge.

\section{EconAgent}\label{sec::method}

\subsection{Perception Module}
To harness the semantic awareness and real-world knowledge capabilities of LLM, we endow each agent with real-world profiles, including name, age, and job. Names are generated by the LLM and randomly assigned to each agent. The age distribution of agents follows the 2018 U.S. population distribution for ages 18-60~\cite{bureau_data_nodate}. For wage and job assignments, we first adjust the parameters of the hourly wage's Pareto distribution to align the monthly wage distribution with 2018 U.S. economic data and tax brackets~\cite{zheng2022ai}. Furthermore, we prompt the LLM to generate ten job titles for each decile of the monthly wage distribution, mirroring the real-world scenario where different jobs have significant wage differences. Agents are initially assigned jobs based on this monthly wage and their jobs are dynamically adjusted throughout the simulation. If an agent chooses to work in the previous month, the job remains unchanged in the following month. If they are unemployed, an offer, based on the current monthly wage, is randomly presented to them. The generated age distribution, monthly wage distribution, and jobs are provided in the supplementary materials.

In addition, we characterize the entire economic environment in a manner closely mirroring the real world, enabling LLM to thoroughly grasp various economic factors. We integrate variations of key economic variables into the prompts and incorporated typical economic keywords to ensure the LLM could fully perceive dynamics in the economic landscape and employ relevant economic principles in its decision-making. For instance, if the agent chose not to work in the previous month, we would supplement the prompt with, 

\begin{center}
\begin{minipage}{0.92\linewidth}
    \begin{shaded}
    \textit{In the previous month, you became unemployed and had no income. Now, you are invited to work as a(an) [offer] with a monthly salary of [wage].}
    \end{shaded}
\end{minipage}
\end{center}
In the prompt, \textit{offer} and \textit{wage} are dynamically adjusted along with the simulation. Such prompting enables the agent to recognize the risks associated with `unemployment', thereby increasing its inclination to work in the subsequent month. More similar designs of prompting are also considered, such as `shortage of goods' when the demand for agents can not be met. 

The perception module enables agents to act as heterogeneous households in the real economic environment, contributing to the emergence of macroeconomic phenomena.

\subsection{Memory Module}
Considering that decision-making within the economic environment is a sequential task, wherein past experiences and economic dynamics play pivotal roles in present decisions, the incorporation of a memory module can assist the agent in fully accounting for market dynamics and in acquiring valuable decision-making insights. Specifically, we dynamically maintain the memory pool with $2L+1$ conversations, encompassing the economic environment and agent decisions from the previous $L$ months. Besides, at the end of each quarter, we input dialogues of this quarter into the LLM, prompting it to `reflect' on the economic phenomena and to respond to how these phenomena might influence its subsequent decisions. The prompts we design are as follows,

\begin{center}
\begin{minipage}{0.92\linewidth}
\begin{shaded}
\textit{Given the previous quarter's economic environment, reflect on the labor, consumption, and financial markets, as well as their dynamics. What conclusions have you drawn?}
\end{shaded}
\end{minipage}
\end{center}
The following is an example of the reflection.
\begin{center}
\begin{minipage}{0.92\linewidth}
\begin{shaded}
\textit{Based on the previous quarter's data, the \textbf{labor market experienced deflation}\ldots The consumption market also saw a \textbf{decrease in prices for essential goods}\ldots The financial market's interest rates \textbf{remained unchanged at 3.00\%}. Overall, the quarter highlighted the need for \textbf{careful financial planning and adaptability in response to market fluctuations}}.
\end{shaded}
\end{minipage}
\end{center}
Obviously, after reflection, agents can fully perceive past market dynamics and adaptively adjust their strategies to maintain daily life and cope with future uncertainties.

The memory module allows the agent to comprehend dynamics in the market and glean reflections from past experiences, modeling the influence of macroeconomic trends.

\subsection{Action Module}
When prompting the LLM for decision-making, we explicitly incorporated considerations of living costs, future aspirations, and economic trends into the prompts. We prompt the LLM to respond with a value in the range [0, 1] to indicate the propensity of working and consumption. 

The action module empowers the agent to automatically account for the influence of various economic factors when making decisions, such as income and savings, leveraging the semantic perception capabilities of LLM. It only requires the inclusion of relevant economic variables in the prompts, without the need for specially designed decision rules.

\section{Experiments}\label{sec::exp}
In this section, we conduct experiments to study the ability of EconAgent, aiming to answer the following research questions (RQ).
\begin{itemize}[leftmargin=*]
    \item \textbf{RQ1}: How do the EconAgent behave in the simulation environment, compared with the traditional agents?
    \item \textbf{RQ2}: How do the perception and reflection modules in EconAgent affect the simulation results?
    \item \textbf{RQ3}: Does the decision-making mechanism of EconAgent possess interpretability, and can it address the challenges faced by traditional agents?
    \item \textbf{RQ4}: Can the simulation based on EconAgent reflect the impact of external intervention?
\end{itemize}

\subsection{Experimental Setup}
We investigate some phenomena of paramount interest in macroeconomics, including several macroeconomic indicators and two macroeconomic regularities. 
For comparative analysis, we select representative rule-based~\cite{lengnick2013agent,gatti2011macroeconomics} and learning-based~\cite{zheng2022ai} agent models.

\paragraph{Definition of Macroeconomics Indicators} Annual nominal GDP is defined as the sum of $S\times P$ over one year. As for real GDP, we set the first year in the simulation as the reference year and replace $P$ with $P_0$, where $P_0$ is the goods price in the reference year. The definition of the annual (price) inflation rate and the unemployment rate is shown in Eq. \ref{def-indicator}. For wage inflation, the definition is similar to that of price inflation, where the average price is replaced with the average wage across all the agents.

\paragraph{Baselines} We select \textbf{LEN}~\cite{lengnick2013agent} and \textbf{CATS}~\cite{gatti2011macroeconomics} as the baselines because 1) they partially reproduce the aforementioned macroeconomics phenomena within their own (more complex) simulation frameworks, and 2) carefully designed decision rules of working and consumption are representative, reflecting typical decision-making observed in real-life scenarios. Considering the importance of agents' heterogeneity in macroeconomic simulation, we also combine these two baselines as an additional baseline \textbf{Composite}, where each agent randomly adopts one rule of them.
Besides, we select a learning-based method, AI-Economist~\cite{zheng2022ai} \textbf{AI-Eco}, which builds on the assumption of rational decision-making and employs reinforcement learning~(RL)~\cite{arulkumaran2017deep} to maximize the agent's utility.

The details of decision rules in LEN and CATS and the training process for AI-Economist are provided in the supplementary materials.

\paragraph{Simulation Parameters} 
We simulate $N=100$ agents. The productivity is set as $A = 1$ for simplicity. The initial goods price is the average hourly wage across all the agents. For the labor and consumption dynamics, $\alpha_w = 0.05$ and $\alpha_P = 0.10$. For the financial market, $r^n = 0.01$, $\pi^t = 0.02$, $u^n = 0.04$, and $\alpha^{\pi} = \alpha^u = 0.5$. Note that our results and conclusions are not sensitive to these parameters.

Our simulation is implemented with Python. We use GPT-3.5-turbo-0613 provided by OpenAI API as the LLM\footnote{https://platform.openai.com/}. In practice, we find $L=1$ works well in the simulation. Other detailed simulation parameters, crucial for replicability and deeper understanding, are provided in the supplementary materials.

\subsection{Macroeconomic Emergence~(RQ1)}
Our evaluation encapsulates a broad spectrum of macroeconomic indicators and regularities~\cite{lengnick2013agent,gatti2011macroeconomics,dawid2018agent,axtell2022agent}. The performance of EconAgent was compared with that of representative rule-based baselines, as detailed in two referenced works and their combination~\cite{lengnick2013agent,gatti2011macroeconomics}. 

\begin{figure}[t!]
\centering
\includegraphics[width=0.96\linewidth]{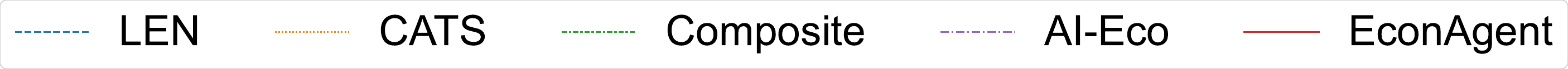} \
\subfloat{\includegraphics[width=0.49\linewidth]{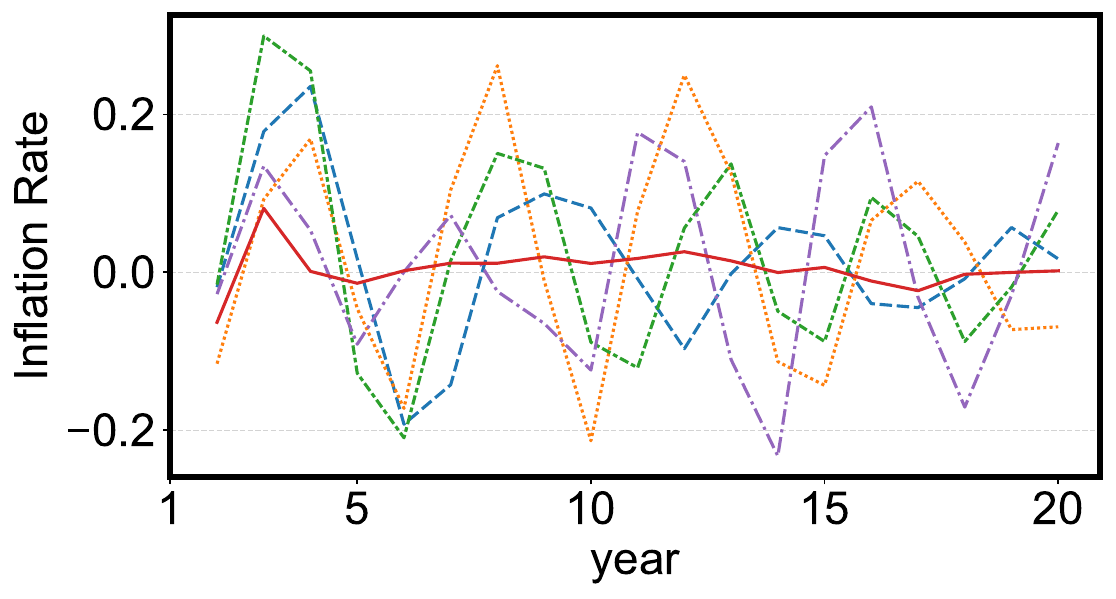}} \subfloat{\includegraphics[width=0.49\linewidth]{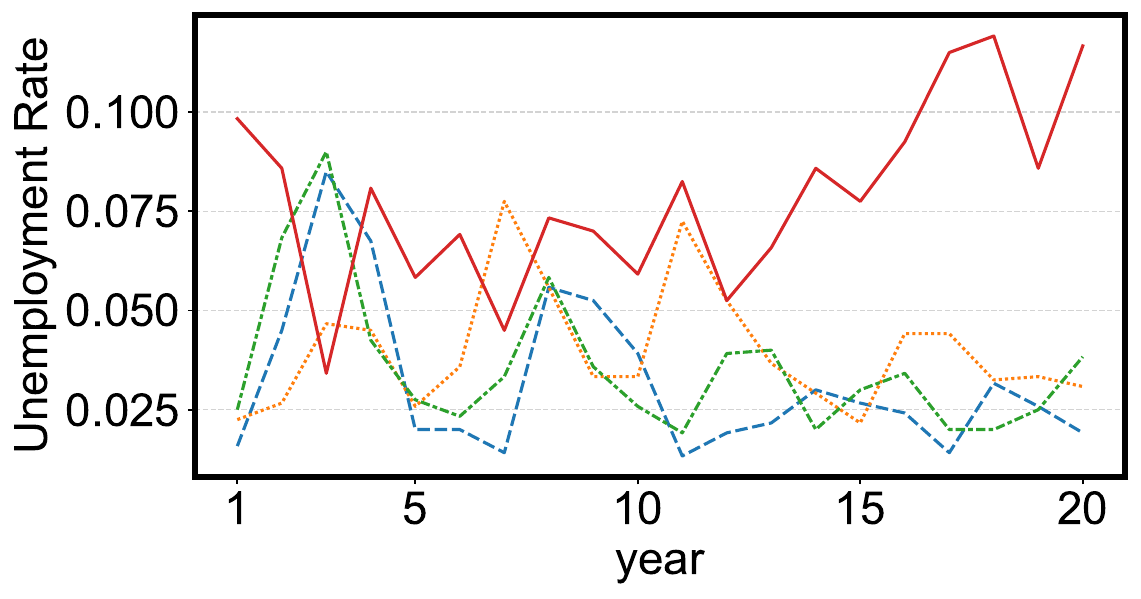}} \
\subfloat{\includegraphics[width=0.49\linewidth]{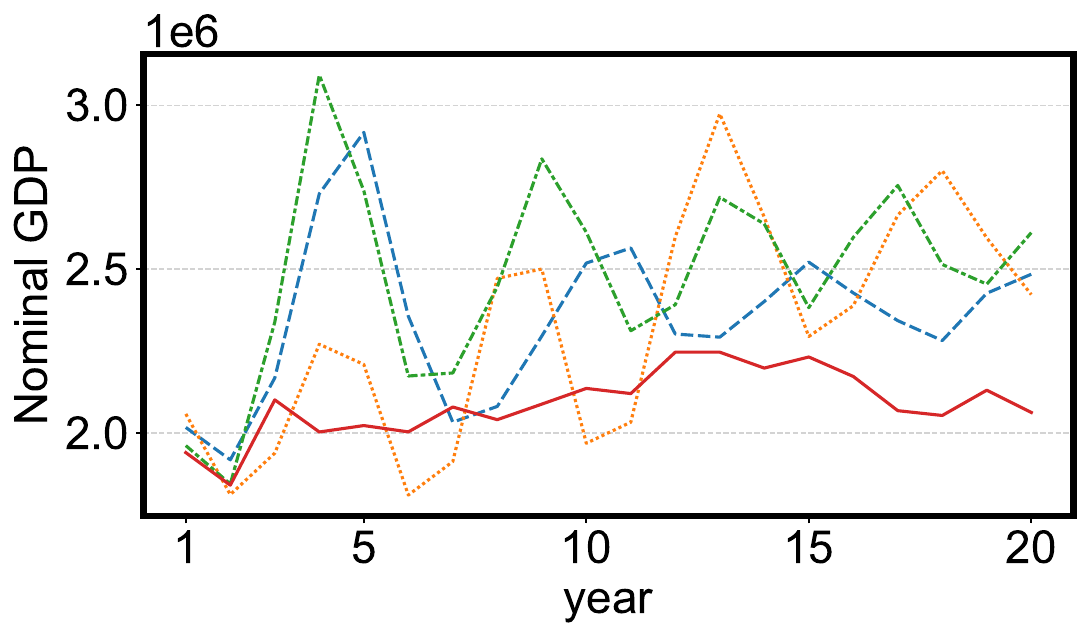}}
\subfloat{\includegraphics[width=0.49\linewidth]{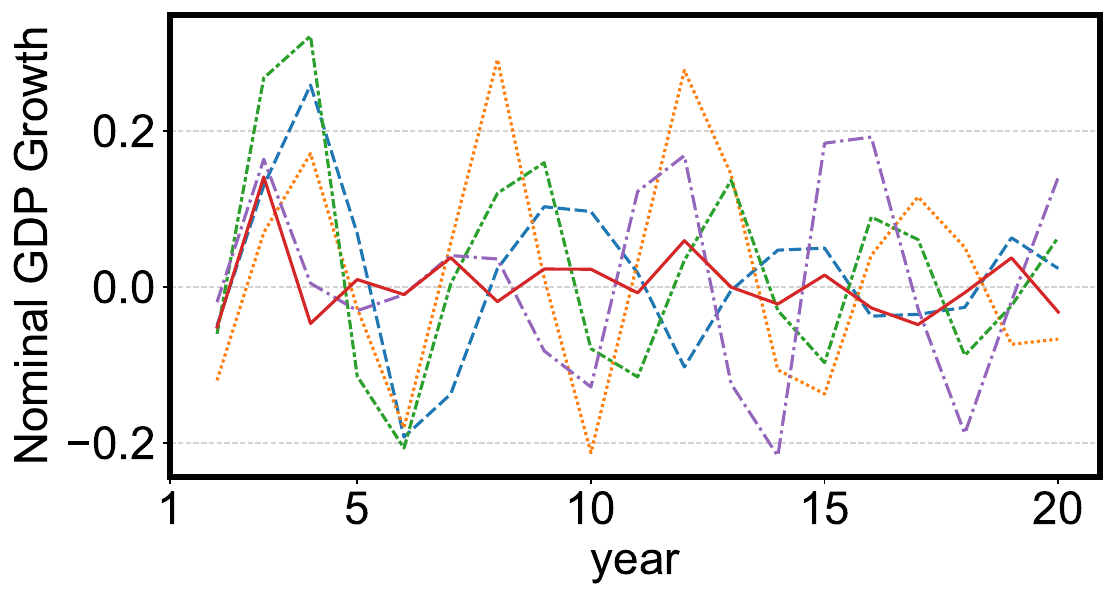}} \
\caption{Annual variations of macroeconomic indicators, where the simulation based on EconAgent shows more stable and numerically plausible indicators.}\label{fig::annual-indicators}
\end{figure}

\paragraph{Macroeconomic Indicators} In Figure~\ref{fig::annual-indicators}, we depict the fluctuations of the annual inflation rate, unemployment rate, nominal GDP, and growth rate of nominal GDP. Note that the unreasonable unemployment rate~(around 46\%) and nominal GDP for AI-Eco are not reported. Both rule-based and RL-driven baselines produce \textbf{anomalous indicators and large fluctuations}. In contrast, agent decision-making based on EconAgent has manifested more \textbf{stable and numerically plausible} macroeconomic phenomena across multiple facets, even without fine-grained calibration. Specifically, the inflation rate consistently fluctuated within a -5\% to 5\% range after the 3-th year, whereas the baselines exhibited significantly larger oscillations, at times even surpassing 20\%. This indicates that the decision-making of EconAgent is coherent and more closely emulates real-world human choices, leading to easier attainment of equilibrium between supply and demand in the consumption market. The unemployment rate generally varied between 2\% and 12\%\footnote{The increase in the unemployment rate after year 15 is actually a normal fluctuation. See Figure \ref{fig::unemployment-c}.}, which aligns well with real-world economic activities~\cite{gatti2011macroeconomics}. Both the nominal GDP and its growth rate also fluctuated within more reasonable numerical bounds like the inflation rate does. We also provide quarterly indicators in the appendix.

\begin{figure}[t]
\centering
\begin{subfigure}{0.96\columnwidth}
    \includegraphics[width=\linewidth]{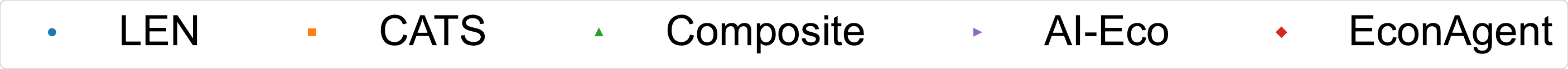}
\end{subfigure}

\includegraphics[width=0.48\columnwidth]{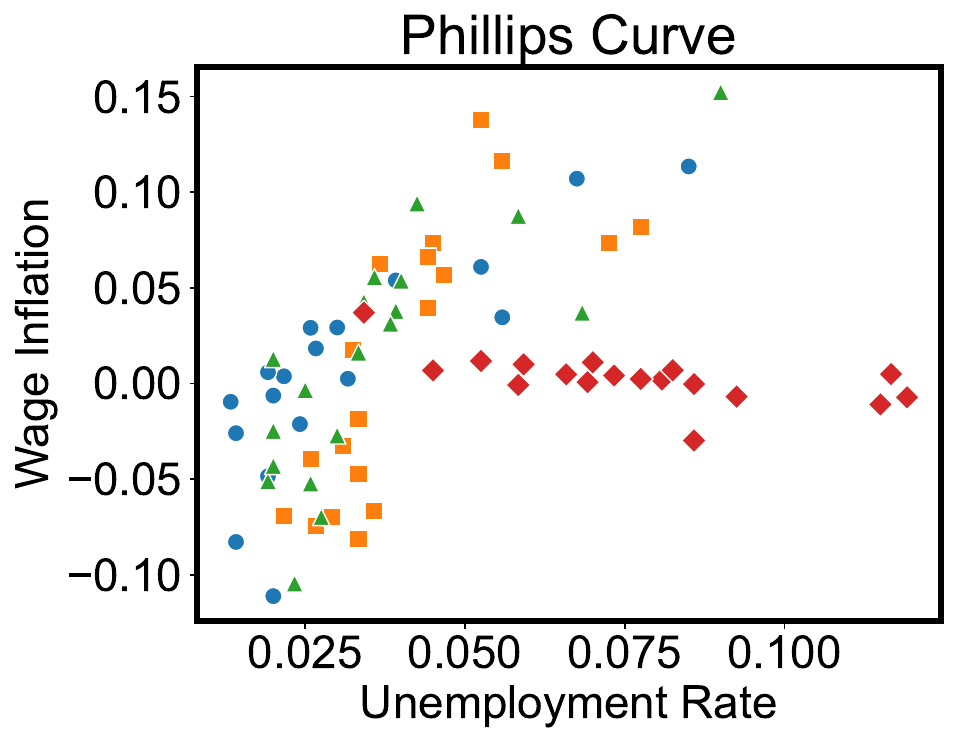}
\includegraphics[width=0.49\columnwidth]{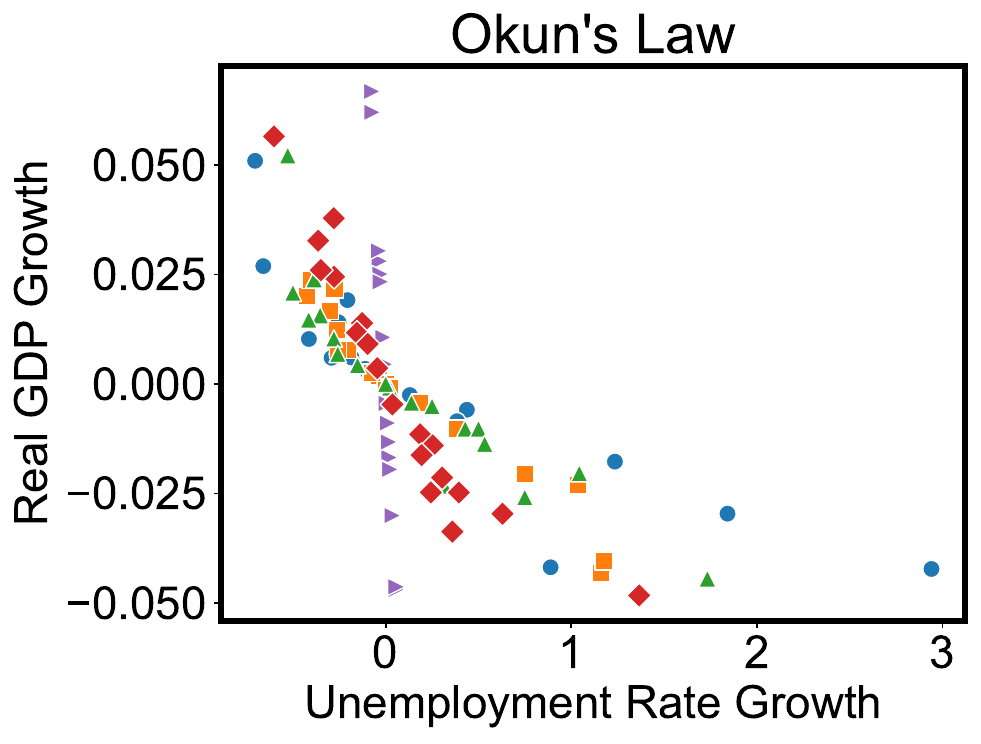}
\caption{Only the simulation based on EconAgent emerges correct Phillips Curve and Okun's Law.} \label{fig::correlation-curves}
\end{figure}

\paragraph{Macroeconomic Regularities} As the two most commonly used regularities in macroeconomic simulations for validating the plausibility of simulation results, the Phillips Curve~\cite{phelps1967phillips} and Okun's Law~\cite{okun1963potential} respectively describe the negative correlations between the annual unemployment rate and wage inflation, and the annual growth rate of unemployment and real GDP growth. As shown in Figure~\ref{fig::correlation-curves}, only the decision-making of EconAgent has \textbf{correctly manifested phenomena in accordance with these two regularities}~(Pearson correlation coefficient is -0.619, $p<0.01$ and -0.918, $p<0.001$). Notably, the rule-based baseline method displayed an \textbf{incorrect positive relationship on the Phillips Curve}. We attribute this advantage to the LLM's accurate perception that consumption should be reduced when unemployed, a point which will be elaborated upon in the subsequent section. Note that the Phillips Curve for AI-Eco is not shown due to the very large unemployment rate.

\subsection{Ablation Study~(RQ2)} 
\begin{figure}[t]
\centering
{\includegraphics[width=0.8\linewidth]{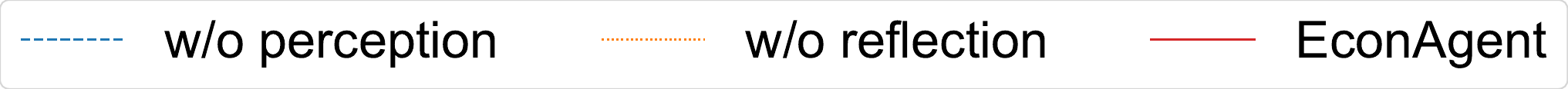}}
\subfloat[Inflation Rate]{\includegraphics[width=0.49\linewidth]{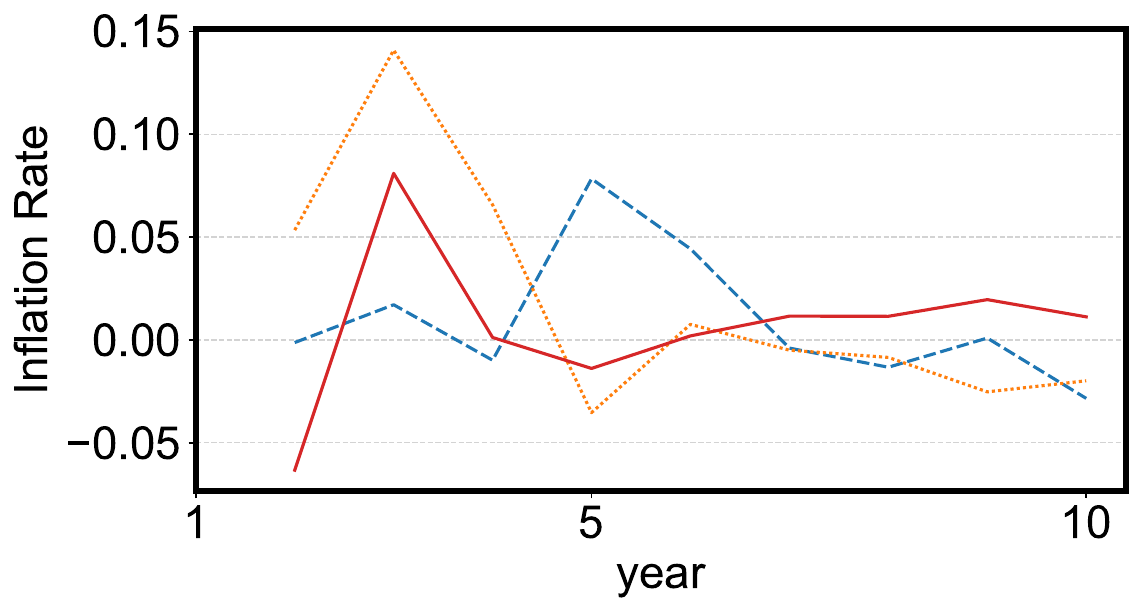}} \
\subfloat[Unemployment Rate]{\includegraphics[width=0.49\linewidth]{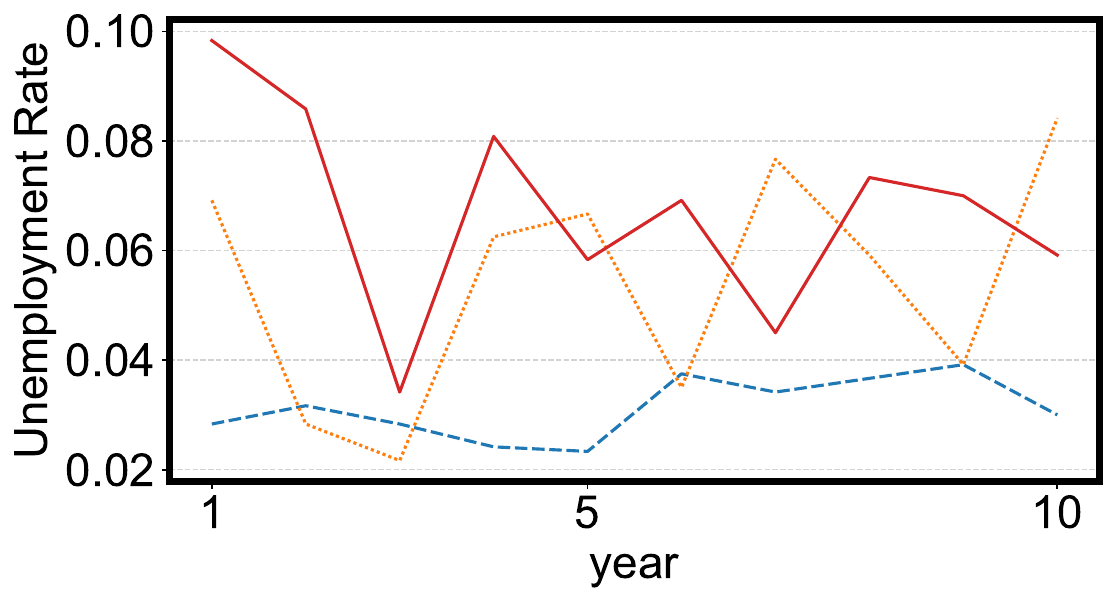}} \
\caption{Ablation study on perception and memory modules.}\label{fig::ablation}
\end{figure}

We separately remove the perception module and the reflection module, and the results of 10 years are as shown in Figure \ref{fig::ablation}. We observe that when there is no perception capability, the inflation rate and unemployment rate fluctuations significantly decrease, appearing "too stable", especially for the unemployment rate. This suggests that the agents have low sensitivity to changes in their economic conditions and cannot make adaptive decision adjustments. When there is no reflection capability, the inflation rate exhibits anomalies close to 15\% in the first three years, emphasizing the importance of long-term (a quarter in our experiments) economic environment perception.

\subsection{Decision-Making Abilities~(RQ3)}
\paragraph{Decision Rationality}
Given that an agent's decisions (propensity of working and consumption) may be influenced by multiple economic variables such as income and savings, we employ regression analysis to delve into the factors affecting these decisions. Specifically, the regression equation is as follows:
\begin{equation}
    p^w_i, p^c_i \sim v_i + \hat{c}_i + T(z_i) + z^r + P + s_i + r,
\end{equation}
where $\hat{c}_i, T(z_i), z^r$ denotes real consumption, the tax paid, and redistribution in the previous month. These independent variables are embedded in the LLM prompts to influence the agent's decisions. We have conducted individual regression analyses on all $N=100$ agents' 240 decisions (spanning 20 years) after applying z-score normalization on all the variables and tabulated the significance of the impact of each economic variable on their decisions. Table \ref{table:num_agents} presents the number of agents for whom the effects of the variables are statistically significant, \textit{i.e.}, $p \leq 0.05$. We observe that 1) the effects of taxation, redistribution, and expected monthly income on the work propensity of the majority of agents are significant, 2) in comparison to work propensity, the previous month's consumption, current savings, and bank interest rates significantly influence the consumption propensity of a greater number of agents, and 3) goods price has a significant impact on both work and consumption propensity for most agents. These phenomena are \textbf{consistent with economic common sense about how people make decisions} in daily life. 

\begin{table}[t]
\caption{The number of agents for whom the effects of the variables on decisions are statistically significant.}
    \centering
    \begin{tabular}{c|c|c|c|c|c|c|c}
    \toprule
     & $v_i$ & $\hat{c}_i$ & $T(z_i)$ & $z^r$ & $P$ & $s_i$ & $r$\\
    \hline
    $p^w_i$ & 60 & 37 & 60	& 65 & 58& 56& 31 \\
    $p^c_i$ & 65 & 73 & 51 & 52 & 62 & 100 & 49 \\
    \bottomrule
    \end{tabular}
    \label{table:num_agents}
\end{table}

\begin{figure}[t]
\centering
\subfloat[Consumption \textit{v.s.} Age]{\includegraphics[width=0.49\linewidth]{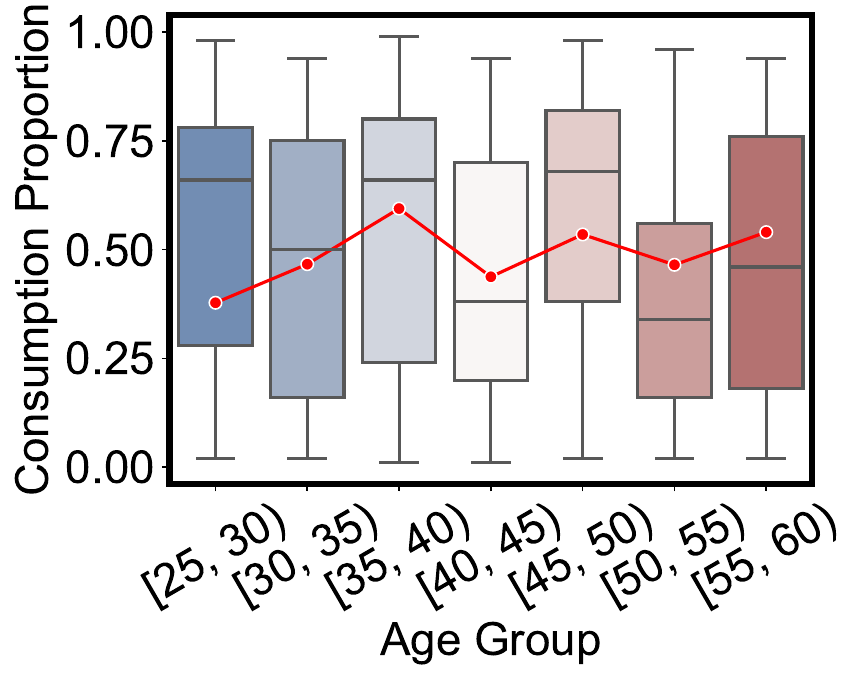}} \
\subfloat[Consumption \textit{v.s.} Unemployment]
{\includegraphics[width=0.47\linewidth]{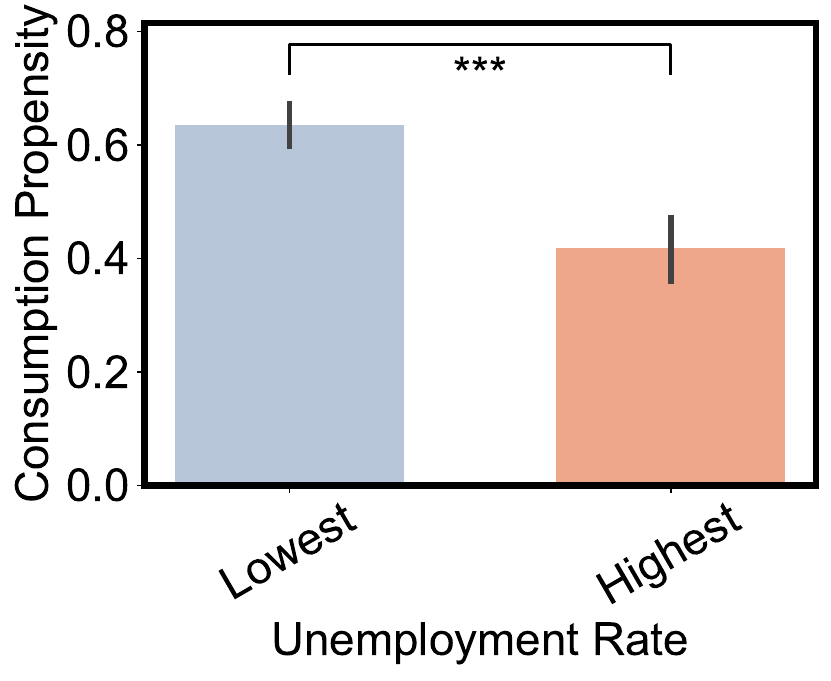}} \

\caption{The modeling of agent heterogeneity and the influence of macroeconomic trends.} \label{fig::explain-Phillips}
\end{figure}

\paragraph{Agent Heterogeneity}
We further investigate whether EconAgent autonomously exhibit heterogeneity in decision-making mechanisms. Figure \ref{fig::explain-Phillips} (a) demonstrates an incremental growth in consumption propensity with age, aligning broadly with empirical regularities observed in traditional macroeconomic research~\cite{carroll1997buffer}.

\paragraph{Influence of Macroeconomic Trends}
Through interactions with the model, we decipher the underlying reasons for the emergence of negative correlations in the Phillips curve. We first calculate the average consumption propensity across all the agents of two years with the highest and lowest unemployment rates. Figure \ref{fig::explain-Phillips} (b) shows the comparison results, where *** denotes a significant disparity with $p<0.001$. Obviously, high unemployment leads to low consumption propensity significantly. To delve deeper into the reasons why agents opt to reduce consumption in the labor market of high unemployment rates, we randomly select an agent and input its conversations with LLM from the year with the highest unemployment rate back into the LLM. We then prompt the LLM to summarize the economic dynamics for each quarter and provide rationales for the consumption decisions made. The following responses demonstrate that agents have a \textbf{perception of macroeconomic trends and are cautious about their spending when facing deflation in the labor market under a high unemployment rate}.

\begin{center}
\begin{minipage}{0.92\linewidth}
\begin{shaded}
\textit{In the last quarter, I have adjusted my \textbf{willingness to work} and my \textbf{planned expenditures} on essential goods slightly \textbf{downwards}. This decision is primarily influenced by the \textbf{continued deflation in the labor market}, resulting in a decrease in my expected income. With a lower income, I need to be \textbf{cautious about my spending} and ensure that I have enough savings for \textbf{future expenses and unforeseen circumstances} \ldots}
\end{shaded}
\end{minipage}
\end{center}

\subsection{External Intervention~(RQ4)}
We further investigate the impact of external interventions on agent behavior and the resulting changes in macroeconomic simulation outcomes, a topic widely discussed in many economic ABM studies~\cite{dawid2018agent}. Using the example of COVID-19, which has had a significant impact on the global economy, we incorporate it into our simulations and conduct comparative experiments.

Modeling the impact of COVID-19 can be conveniently implemented by incorporating them into the prompt of a EconAgent. Specifically, we add the following statement to the prompt for simulations after March 2020:
\begin{center}
\begin{minipage}{0.92\linewidth}
\begin{shaded}
\textit{In response to the large-scale outbreak of COVID-19 in the United States, the federal government has declared a national emergency since March 2020.}
\end{shaded}
\end{minipage}
\end{center}

\begin{figure}[t]
\centering
\includegraphics[width=0.8\linewidth]{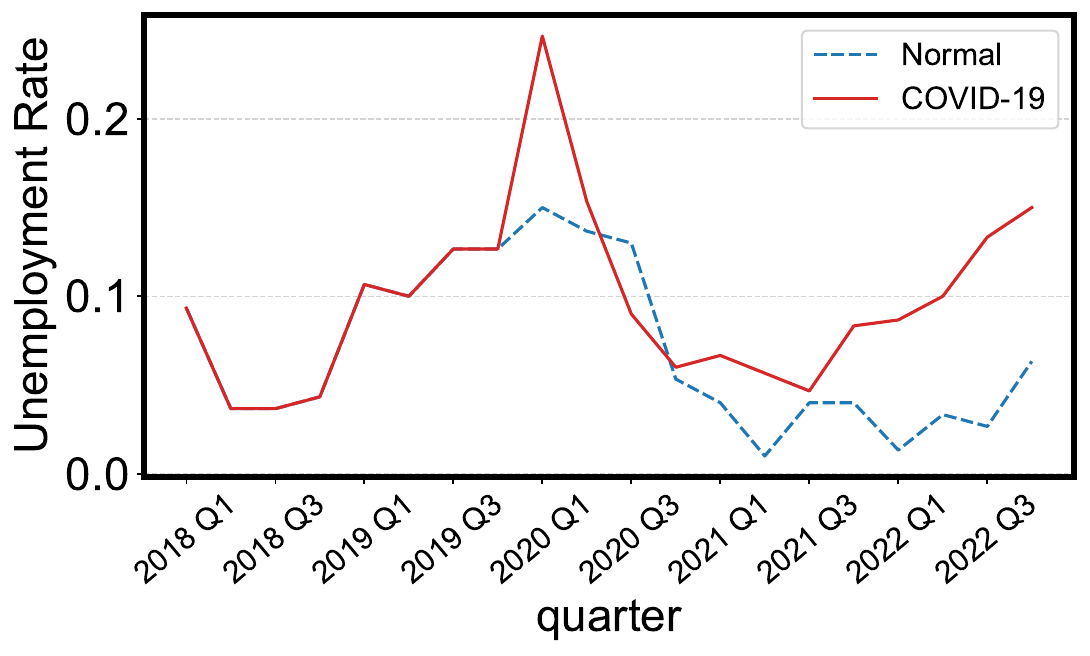}
\caption{COVID-19 brings the surge of simulated unemployment rates.} \label{fig::covid}
\end{figure}

Figure \ref{fig::covid} shows the comparison of unemployment rates, where `Normal' and `COVID-19' denote the simulation results w/ and w/o the prompt above, respectively. This indicates that the simulation based on the EconAgent successfully replicated the surge in the unemployment rate in 2020 Q1 under the impact of COVID-19~\cite{organization_for_economic}. Although the numerical values do not match the real data exactly, this demonstrates that our proposed framework possesses the ability to qualitatively simulate human decision-making and macroeconomic phenomena in the real world. Furthermore, since we did not introduce government intervention measures, the unemployment rate after 2021 remains significantly higher than in the 'Normal' scenario, reflecting the lasting impact of COVID-19.

The following is an example of the agent's reflection during COVID-19, demonstrating its human-like decisions.
\begin{center}
\begin{minipage}{0.92\linewidth}
\begin{shaded}
\textit{\ldots However, the outbreak of COVID-19 and the subsequent national emergency declaration had a significant impact on the labor market. \ldots resulting in widespread unemployment and uncertainty. This situation has likely affected my willingness to work, as job security and health concerns become more prominent \ldots}
\end{shaded}
\end{minipage}
\end{center}
\section{Related Work}
\label{sec::related}

\subsection{Simulation in Macroeconomics}\label{sec::relatedwork1}

ABM emerged as a more promising solution for macroeconomic research compared with empirical statistical models~\cite{hendry1982formulation,phelps1967phillips,kydland1982time} and DSGE models~\cite{christiano2005nominal} in recent decades.
Diverse agents interact with each other based on rules or computational models, avoiding the assumption of a predetermined economic equilibrium. These models allow for a wide range of nonlinear behaviors, enabling policymakers to simulate different policy scenarios and qualitatively assess their impacts on the economy.
The agent modeling with predetermined rules~\cite{tesfatsion2006handbook,brock1998heterogeneous} or neural networks~\cite{trott2021building,zheng2022ai,mi2023taxai} still suffers from the assumption of oversimplified agent behavior or the dependence on large-scale training data, thus having limitations in capturing the full complexity of economic systems.

In this work, we introduce EconAgent with reasoning and planning abilities for simulating macroeconomic activities.

\subsection{LLM-empowered Agents}\label{sec::relatedwork1}
Recently, LLMs trained with large-scale corpus have shown human-level abilities and provide the basis for constructing agents for simulation~\cite{wang2023survey,xi2023rise}.
The LLM agents primarily have three advantages for simulation, including autonomous but adaptive reactions~\cite{autogpt, babyagi}, human-like intelligence for planning~\cite{wang2023survey,xi2023rise}, and interaction and communication with other agents or human~\cite{park2023generative,gilbert2005simulation,park2023generative}. With these advanced abilities, LLM agents have been widely used in many areas, including social science~\cite{park2022social,park2023generative,kovavc2023socialai,gao2023s,jinxin2023cgmi}, natural science~\cite{boiko2023emergent,bran2023chemcrow}, \textit{etc}. Moreover, some works also consider LLM agents in economic research, which can be categorized into three levels~\cite{gao2023large}, including rationality or bias in individual behavior~\cite{horton2023large,chen2023emergence}, planning and cooperation in interactive behavior~\cite{guo2023gpt,akata2023playing}, and system-level market simulation~\cite{zhao2023competeai,anonymous2024rethinking,chen2023put}. 

In short, existing works only consider individual one-step or few-step behavior for a few agents, without experimenting on multi-step behaviors within a multi-agent simulation environment, which is the focus of our work.

\section{Conclusion}\label{sec::conclusion}
In this work, we ventured into the novel integration of LLMs with macroeconomic simulation, designing EconAgent with the abilities of perception, reflection, and decision-making based on the context of real-world economic environments. Classic macroeconomic phenomena are reproduced and more reasonable compared to traditional rule-based or learning-based agents. Through this endeavor, it has become evident that the capabilities of LLMs offer a promising avenue to simulate more realistic macroeconomics.

\section{Limitations}
\paragraph{Simulation Environment} Although our constructed simulation environment encompasses a variety of market dynamics, it currently only simulates the labor and consumption behavior of households. More complex agents (\textit{e.g.}, firms) and their behaviors (\textit{e.g.}, pricing, hiring) have not yet been incorporated. Simulating these agents with LLMs can facilitate a move towards a more realistic macroeconomic system, giving rise to more complex empirical regularities, such as the Beveridge curve and procyclicality/countercyclicality~\cite{dawid2018agent}.

\paragraph{Optimization and Forecasting} Current results are confined to replicating stylized facts~\cite{dawid2018agent}. However, another primary goal of macroeconomic research is policy optimization, along with accurate forecasts of key macroeconomic indicators~\cite{poledna2023economic}. Therefore, EconAgent still requires more realistic behavioral responses to subtle changes in macroeconomic policies. Moreover, constructing a macroeconomic system that closely mirrors the real world necessitates the implementation of EconAgent at the urban scale level, which cannot be achieved without further breakthroughs in LLMs, including enhancements in inference speed and reductions in computational demands.

\section*{Acknowledgement}
This work is supported by the National Natural Science Foundation of China under U23B2030 and 72342032.

\clearpage
\bibliography{bibliography}

\appendix
\appendix
\section{EconAgent}
\begin{figure}[h]
\centering
\includegraphics[width=0.44\columnwidth]{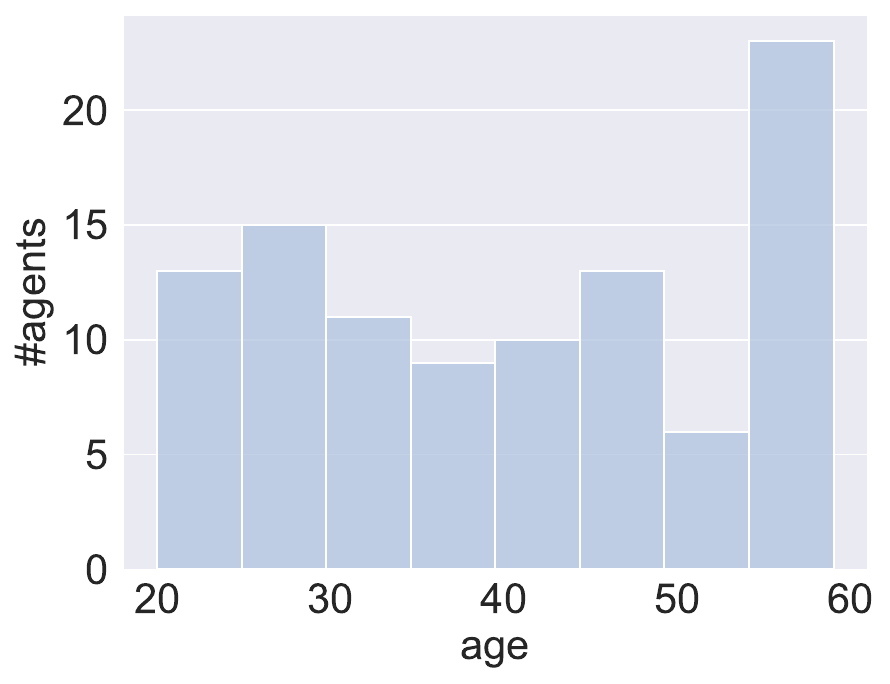}
\includegraphics[width=0.49\columnwidth]{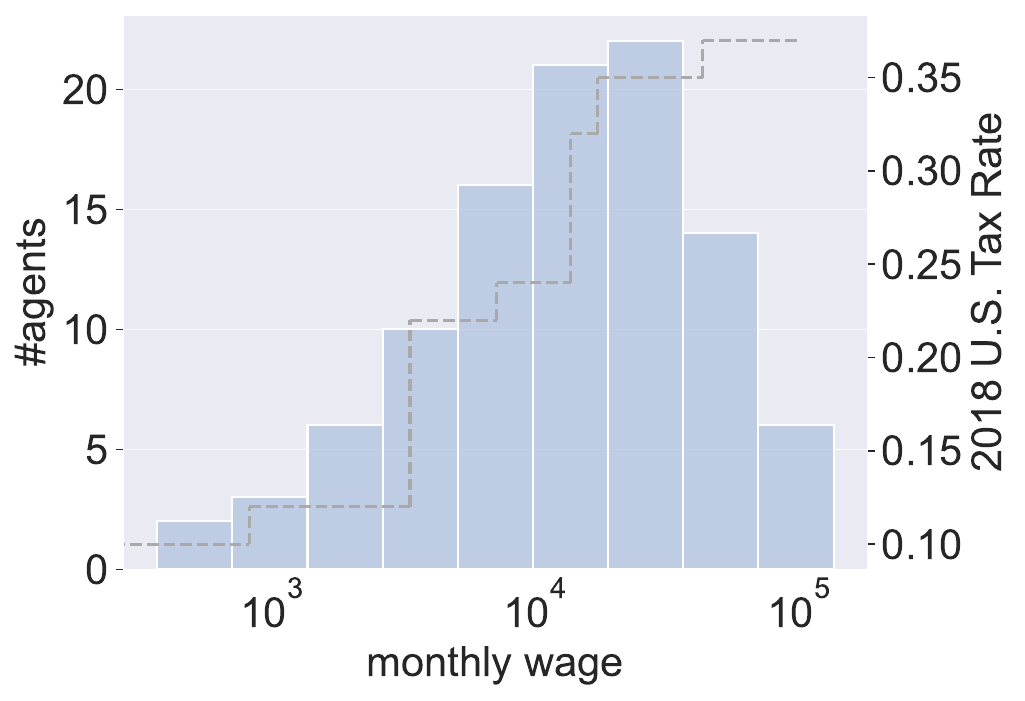}
\caption{Age and monthly wage distribution for agent profiles.} \label{fig::age_wage}
\end{figure}

\paragraph{Agent Profiles} The initialization of agents' profiles is shown in Figure \ref{fig::age_wage}, including age distribution~(left) and monthly wage distribution~(right), as well as the tax brackets and rates of U.S. federal government in 2018, represented by the gray dotted line. As for the generated jobs aligned with the monthly wage, we show some examples as follows,
\begin{itemize}[leftmargin=*]
    \item \([0, 2454)\): Dog Walker, House Cleaner, Newspaper Delivery \(\ldots\)
    \item \([2454, 4838)\): Barista, Cashier, Fast Food Worker \(\ldots\)
    \item \([35469, 52370)\): Psychiatrist, Pediatrician, Anesthesiologist \(\ldots\)
\end{itemize}

\paragraph{Economic Prompts} We provide a full prompt to illustrate our consideration of economic factors, as well as other details not mentioned in the main text.

\begin{mdframed}[backgroundcolor=gray!20, skipabove=\baselineskip, skipbelow=\baselineskip]
You're Adam Mills, a 40-year-old individual living in New York City, New York. As with all Americans, a portion of your monthly income is taxed by the federal government. This taxation system is tiered, income is taxed cumulatively within defined brackets, combined with a redistributive policy: after collection, the government evenly redistributes the tax revenue back to all citizens, irrespective of their earnings. Now it's 2001.02. In the previous month, you worked as a(an) Professional Athlete. If you continue working this month, your expected income will be \$84144.58, which is decreased compared to the last month due to the deflation of the labor market. Besides, your consumption was \$49825.69. Your tax deduction amounted to \$28216.98. However, as part of the government's redistribution program, you received a credit of \$6351.29. In this month, the government sets the brackets: [0.00, 808.33, 3289.58, 7016.67, 13393.75, 17008.33, 42525.00] and their corresponding rates: [0.10, 0.12, 0.22, 0.24, 0.32, 0.35, 0.37]. Income earned within each bracket is taxed only at that bracket's rate. Meanwhile, deflation has led to a price decrease in the consumption market, with the average price of essential goods now at \$135.82. Your current savings account balance is \$12456.42. Interest rates, as set by your bank, stand at 3.00\%. With all these factors in play, and considering aspects like your living costs, any future aspirations, and the broader economic trends, how is your willingness to work this month? Furthermore, how would you plan your expenditures on essential goods, keeping in mind goods price? Please share your decisions in a JSON format. The format should have two keys: 'work' (a value between 0 and 1 with intervals of 0.02, indicating the willingness or propensity to work) and 'consumption' (a value between 0 and 1 with intervals of 0.02, indicating the proportion of all your savings and income you intend to spend on essential goods).
\end{mdframed}

\section{Experimental Setup}
\paragraph{Baselines}
\begin{itemize}[leftmargin=*]
    \item \textbf{Consumption}. In LEN, the consumption decision is \textit{memory-based}, which means that consumption is influenced not only by current income but also by past accumulated savings. Besides, the goods price is another important factor. Conversely, in CATS, it is \textit{non-memory-based} consumption decisions suggesting that consumption is solely related to the current income. The agent aims to keep a desired ratio between savings and income, and consumption is only a proportion of the income. For more human-like decisions, we also introduce the influence of interest rate in the decision rule.

    For LEN, the calculation of consumption propensity is as follows,
    \begin{equation}
        p_i^c = \left(\frac{P}{s_i+z_i}\right)^{\beta}, \beta\in [0, 1].
    \end{equation}
    For CATS, the calculation is as follows,
    \textit{i.e.},
    \begin{equation}
        \frac{\hat{s}_i}{z_i} = \frac{(1+r)(s_i + (1-c)z_i)}{z_i} = h, p_i^c = \frac{c z_i}{s_i + z_i},
    \end{equation}
    where $\hat{s}_i$ denotes the expected savings after consumption in the next month, $h$ is a constant, and $c$ indicates the consumption proportion of the current income. Refer to CATS for the calculation of $c$. Note that we introduce the influence of the interest rate $r$ to endow the agent with the perception of fiscal policy.

    \item \textbf{Work}. The rule of working in LEN and CATS can not be directly used in our simulation framework because we don't simulate firms. Therefore, we follow the intuitions of their designs and define a formula implying that a higher expected income, lower savings, or a lower interest rate leads to a greater propensity to work.

    The work propensity is calculated as
    \begin{equation}
        p^w_i = \left(\frac{v_i}{s_i(1+r)}\right)^{\gamma}, \gamma\in [0, 1].
    \end{equation}

\end{itemize}

For AI-Economist, the utility is a satisfaction function positively correlated with savings and consumption but negatively correlated with labor. Maximizing utility implies that the agent always desires more savings and consumption but prefers less labor. The policy network for the agent's work and consumption decisions is a multi-layer perceptron~(MLP), where the input includes various environment information, such as monthly wage, interest rate, goods price, tax rates, \textit{etc}. We modify the utility function to incorporate consumption and goods price, defined as
\begin{equation}
    \frac{(s_i/P)^{1-\lambda_s}-1}{1-\lambda_s}\cdot \frac{(\hat{c}_i/P)^{1-\lambda_c}-1}{1-\lambda_c} - \lambda_l l_i,
\end{equation}
where $\lambda_{s,c,l}$ balance the importance of savings, consumption, and labor contributing to agent satisfaction. Besides, it's discouraged to not work or have no consumption at all, which leads to negative utility. We also introduce the goods price to make the AI agent perceive the dynamics of the consumption market.

\paragraph{Simulation Parameters} For LEN, CATS, and Composite, we conduct a careful grid search for proper hyperparameters $\beta, \gamma, h$ in decision rules, with the search spaces of $[0.05, 0.1, 0.3, 0.5], [0.05, 0.1, 0.3, 0.5]$, $[0.5, 1, 3, 5]$, respectively. The reported results in the main text are obtained with $\beta=0.1, \gamma=0.1, h=1$, which show the most reasonable macroeconomic indicators.

\paragraph{LLM Costs} Each simulation based on EconAgent incurs a cost of approximately \$30 and takes about 2 hours to complete.

\section{Additional Results} 

\begin{figure}[t]
\centering
\includegraphics[width=0.44\columnwidth]{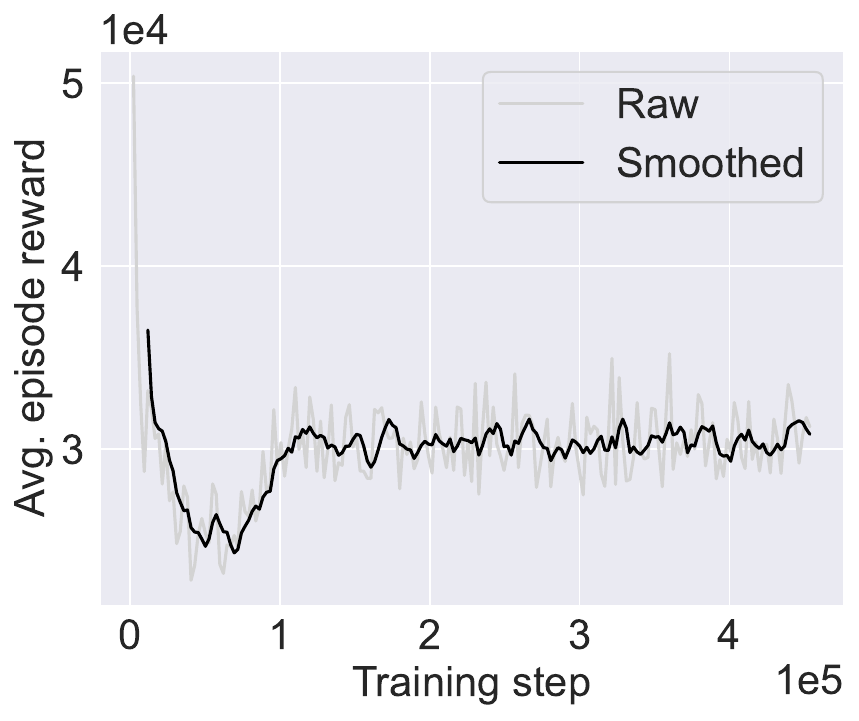}
\includegraphics[width=0.44\columnwidth]{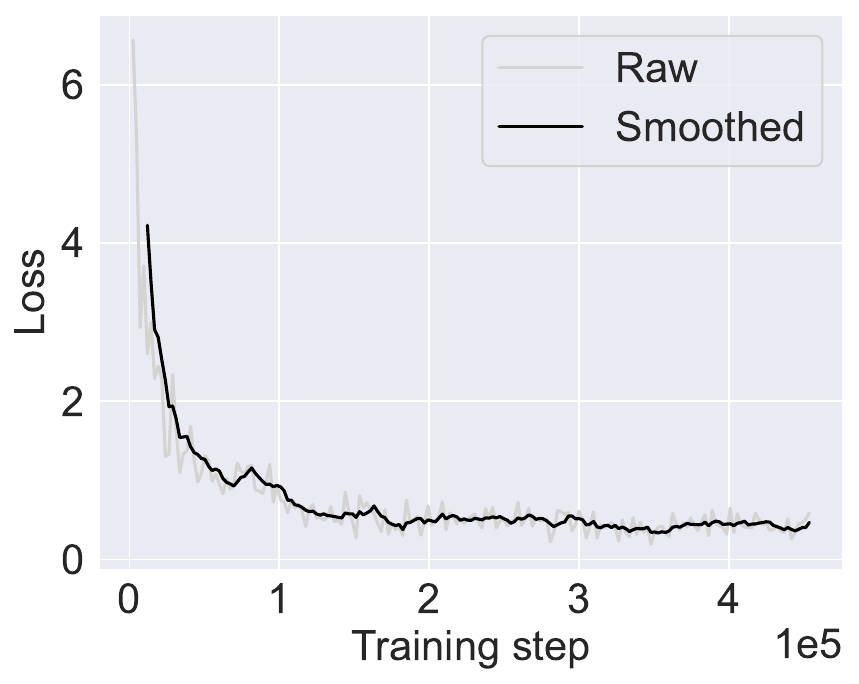}
\caption{The training process of AI-Economist.} \label{fig::ai}
\end{figure}

\begin{figure}[h!]
\centering
\includegraphics[width=0.9\linewidth]{figs/legend-line.pdf} \
\subfloat{\includegraphics[width=0.9\linewidth]{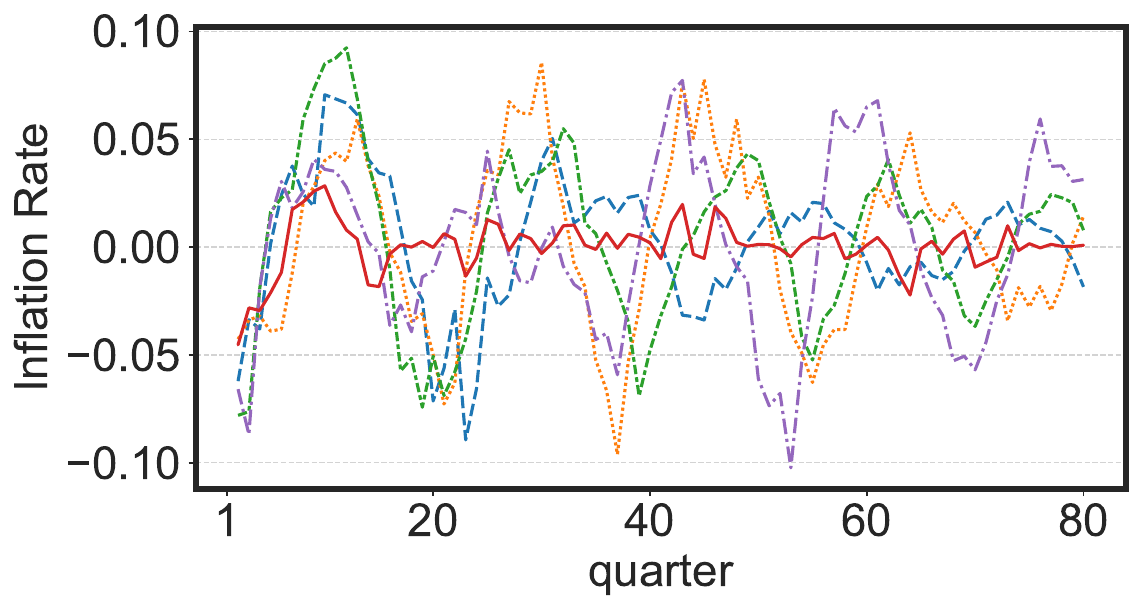}} \
\subfloat{\includegraphics[width=0.9\linewidth]{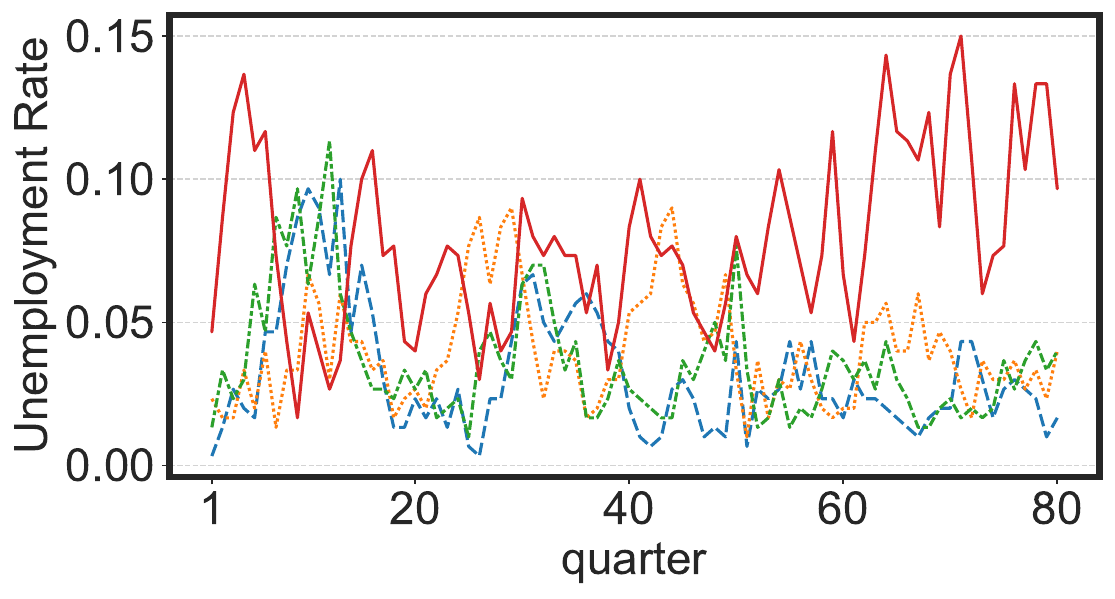}} \
\subfloat{\includegraphics[width=0.9\linewidth]{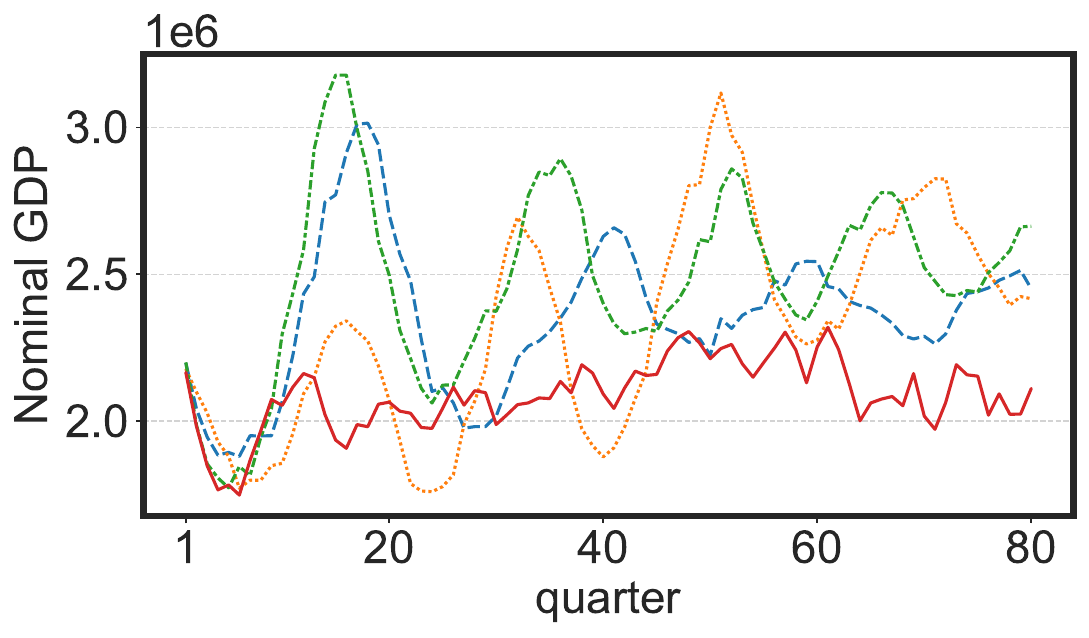}} \
\subfloat{\includegraphics[width=0.9\linewidth]{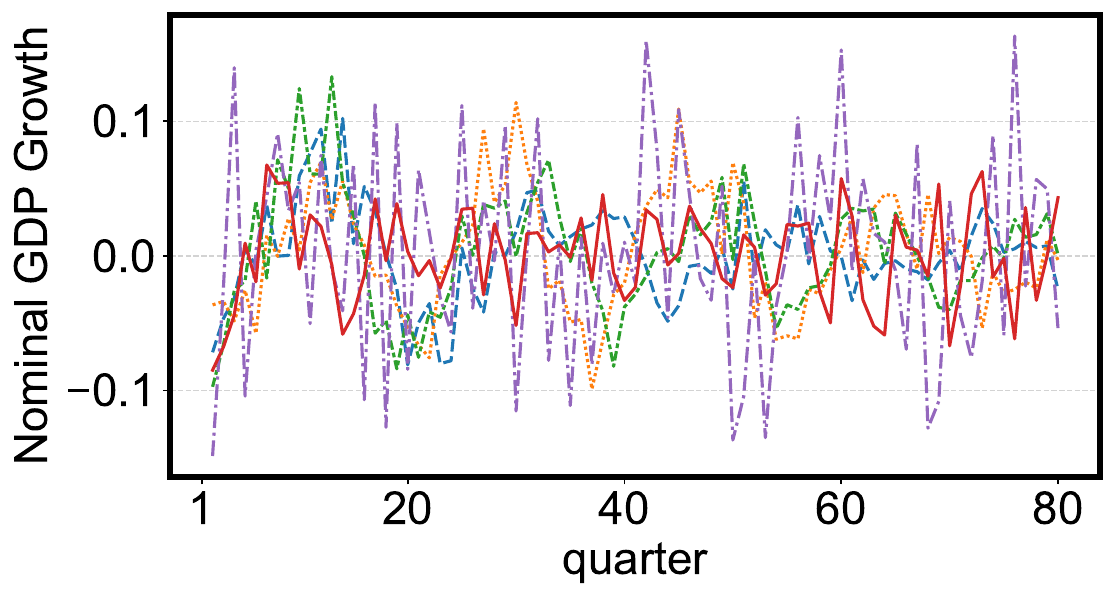}} \
\caption{Quarterly variations of macroeconomic indicators.}\label{fig::quarter-indicators}
\end{figure}

\paragraph{Quaterly Indicators} Figure~\ref{fig::quarter-indicators} presents quarterly macroeconomic indicators, where the conclusion is similar to that of annual ones. For AI-Economist, we follow ~\cite{zheng2022ai} to adopt PPO algorithm~\cite{schulman2017proximal} to train the policy network, where the actor and critic networks have the hidden dimensions of $[128, 128]$ and $[128, 64, 32]$, respectively. The observation~(input) dimension is 173 and the action~(output) dimension is 53, including 2 work actions and 51 consumption actions~(0-1 with an interval of 0.02). The training process is shown in Figure \ref{fig::ai}, including the loss and average episode reward, where one episode is a complete simulation of 20 years.

\begin{figure}[h!]
\centering
\includegraphics[width=0.9\linewidth]{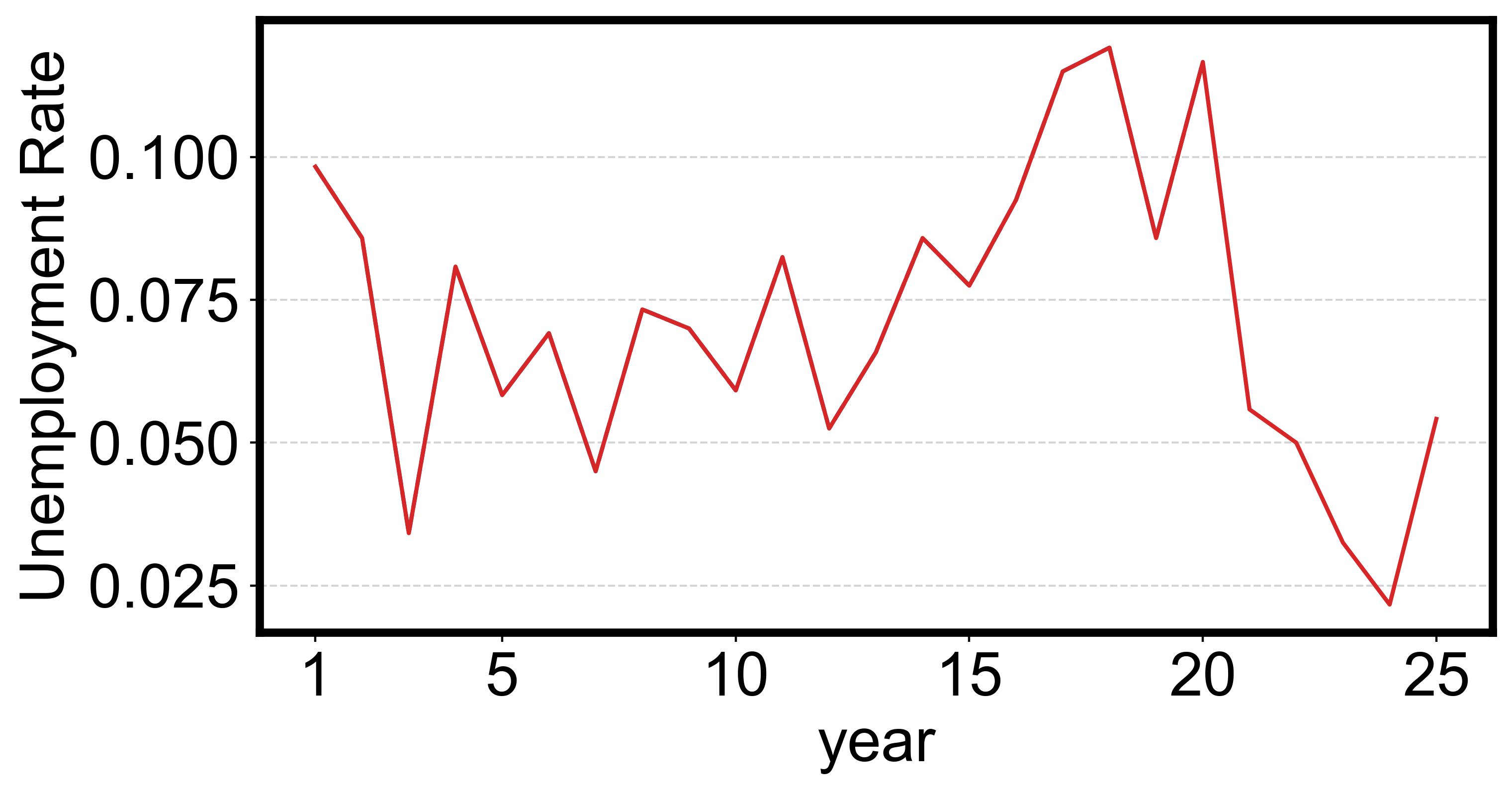} \
\caption{Unemployment rate for 25 years.}\label{fig::unemployment-c}
\end{figure}

\paragraph{Fluctuated Unemployment Rate} The unemployment rate after $5$ additional years of simulation is shown in Figure \ref{fig::unemployment-c}, which returns to a lower level after the 20th year.

\begin{figure}[h!]
\centering
{\includegraphics[width=0.9\linewidth]{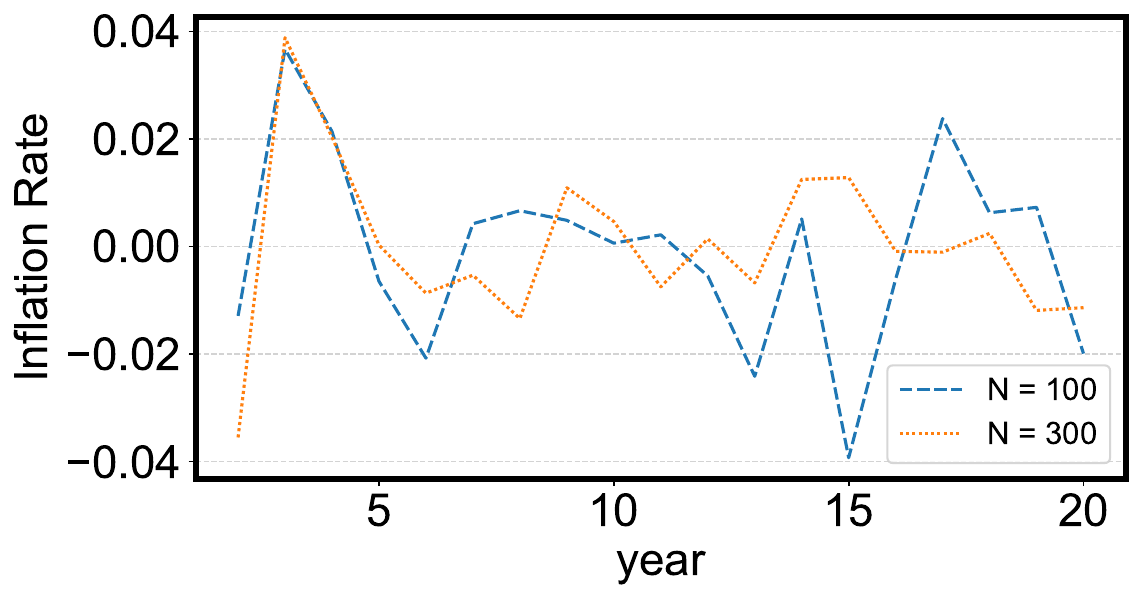}}
\caption{Inflation rate under different number of agents.}\label{fig::sensitivity}
\end{figure}

\paragraph{Sensitivity and Robustness} We increase the number to 300 and run the simulation again. Figure \ref{fig::sensitivity} shows consistently stable and plausible inflation rates similar to those of 100 agents, which holds for other indicators as well. Therefore, the simulation results are insensitive to the number of agents.

\begin{figure}[h!]
\centering
\subfloat{\includegraphics[width=0.9\linewidth]{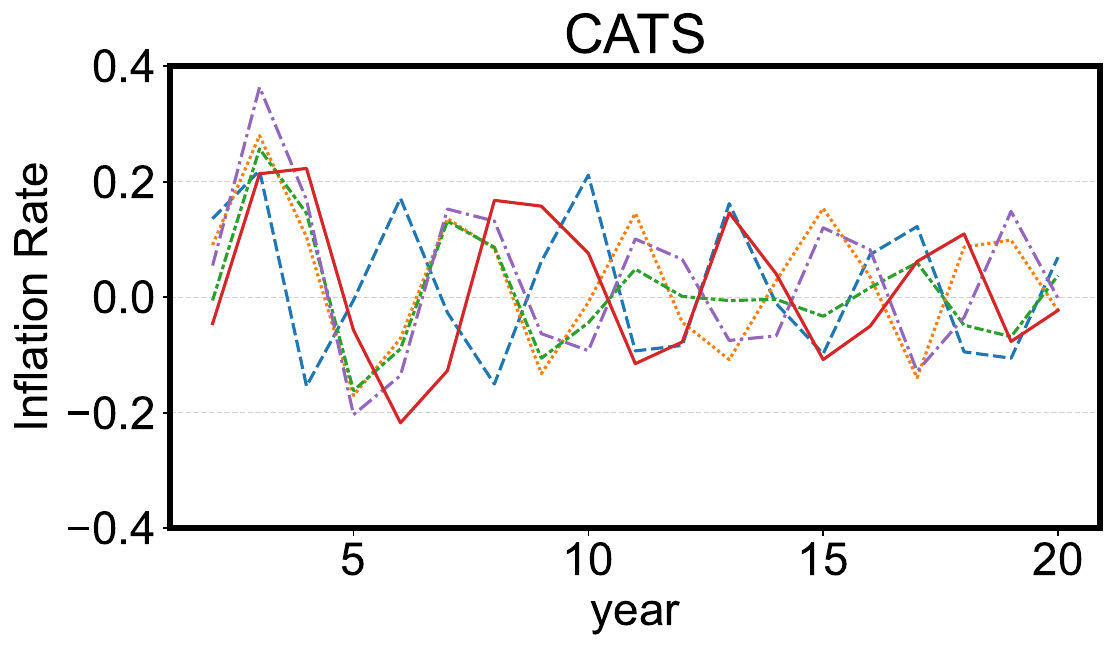}} \
\subfloat{\includegraphics[width=0.9\linewidth]{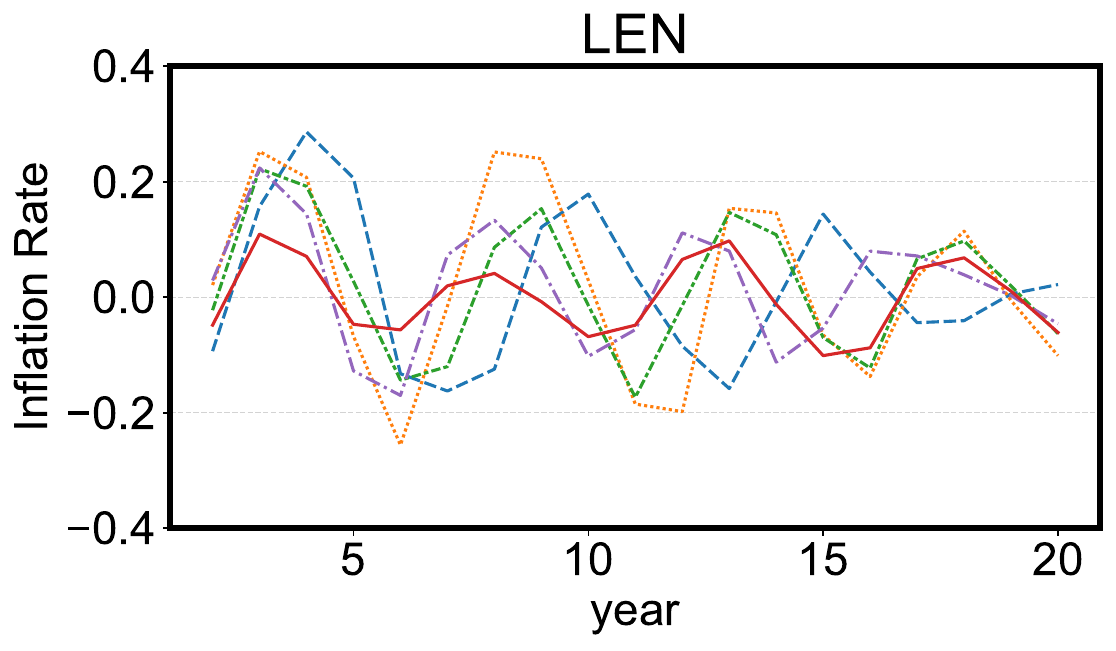}} \
\subfloat{\includegraphics[width=0.9\linewidth]{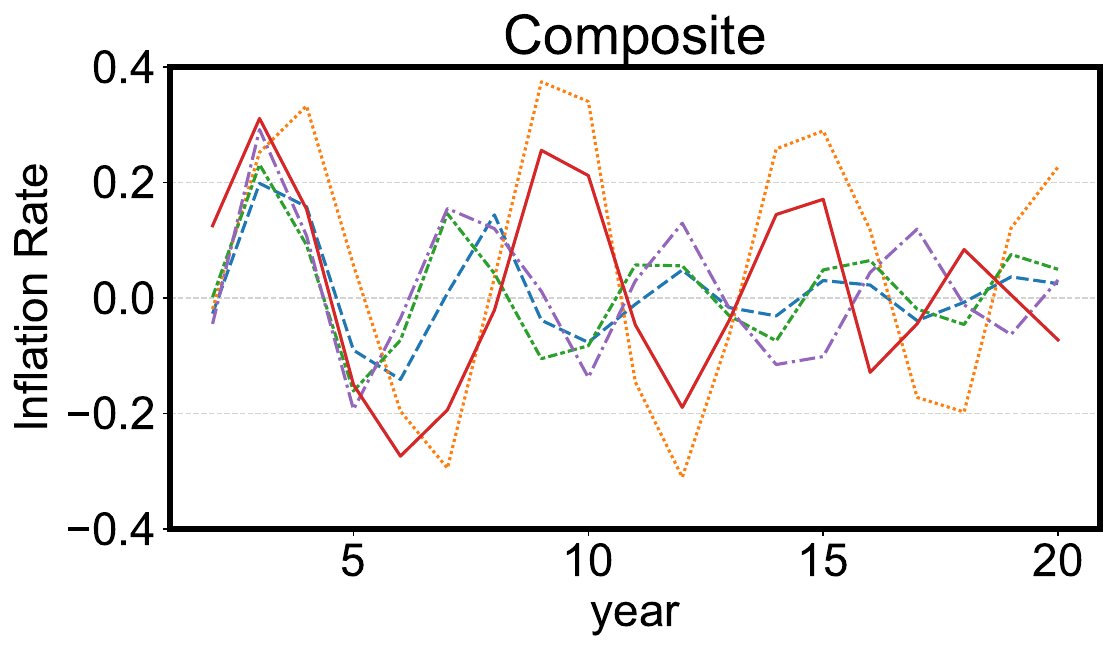}} \
\subfloat{\includegraphics[width=0.9\linewidth]{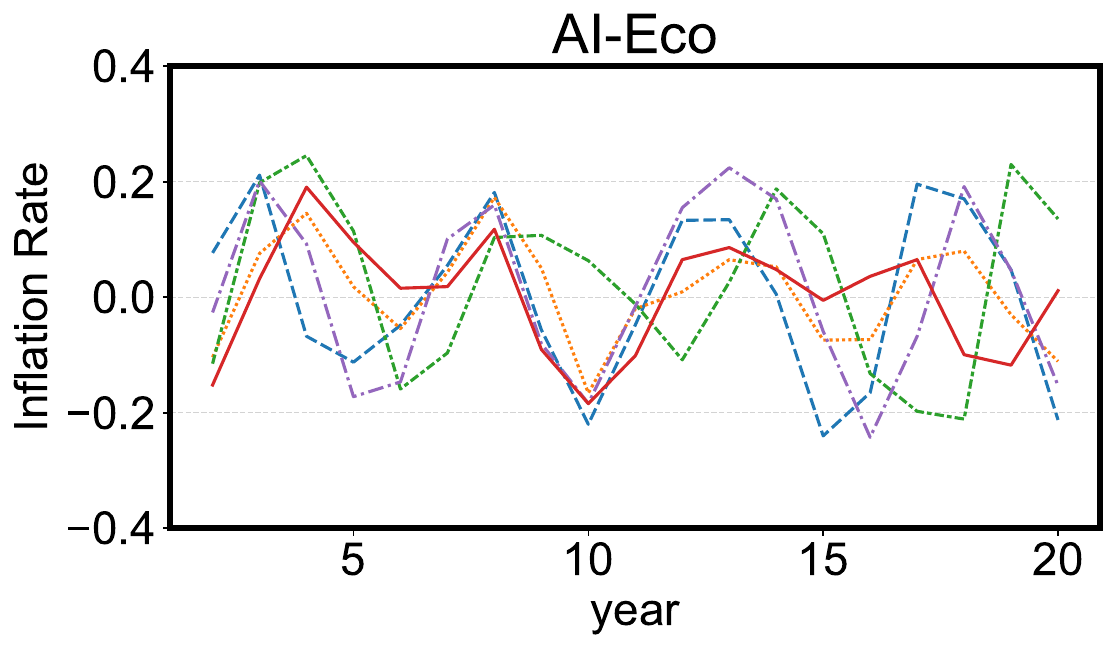}} \
\subfloat{\includegraphics[width=0.9\linewidth]{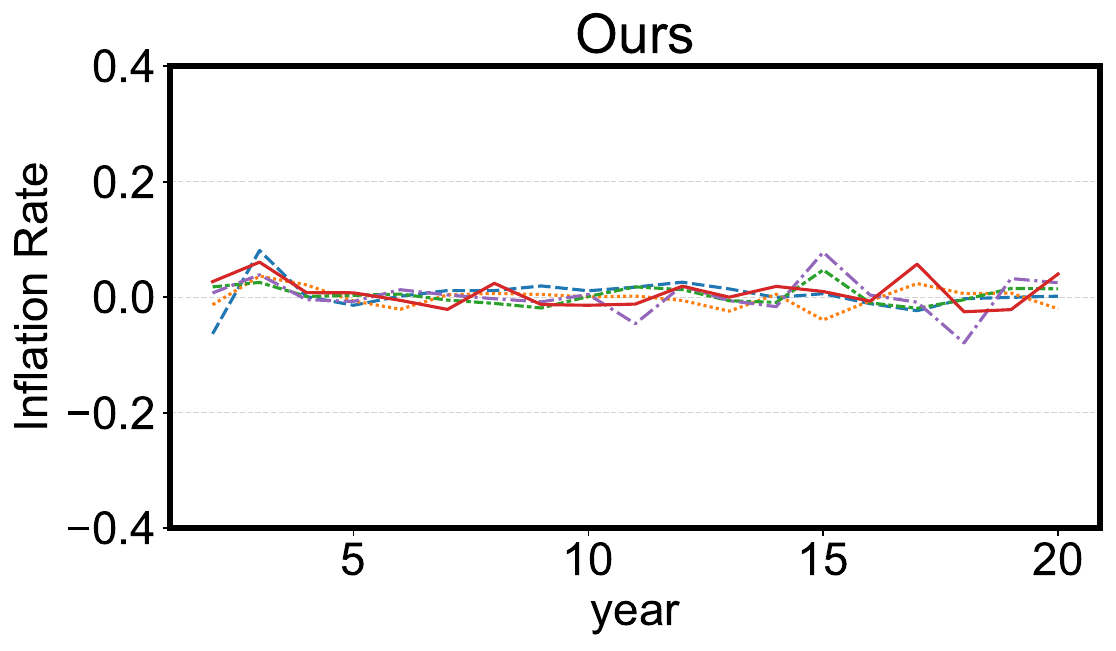}} \
\caption{Simulation variance across five experiments for different agent models.}\label{fig::robustness}
\end{figure}

We conduct five simulations for each agent model and present the inflation rate in Figure \ref{fig::robustness}. It is evident that the simulations based on EconAgent are robust and consistently yield more stable and plausible results, which is also true for other indicators. Importantly, there are no significant differences in terms of stability and numerical scale among the simulations.

\end{document}